\documentclass[sigconf,screen]{acmart}
\usepackage{graphicx}
\usepackage{booktabs}
\usepackage{multirow}
\usepackage{colortbl}

\usepackage{tcolorbox}
\tcbuselibrary{skins, breakable}
\usepackage{amsmath}
\usepackage{graphicx}
\usepackage{booktabs}
\usepackage{csquotes}
\usepackage{pifont}
\usepackage{algorithm}
\usepackage{algpseudocode}
\usepackage{multirow}
\usepackage{xcolor}
\usepackage{amsmath}
\usepackage{tabularx}
\usepackage{array}
\usepackage{fontawesome5}
\definecolor{darkgreen}{rgb}{0.0, 0.5, 0.0}
\definecolor{lorablue}{rgb}{0.1, 0.1, 0.6}
\newcolumntype{C}{>{\centering\arraybackslash}X}

\copyrightyear{2026}
\acmYear{2026}
\setcopyright{none}
\acmConference[Preprint]{Preprint}{2026}{}
\acmBooktitle{Preprint}
\renewcommand\footnotetextcopyrightpermission[1]{}

\renewcommand{\shortauthors}{Ding et al.}

\title{CausalSplat: Towards Comprehensive Hierarchical Reasoning in 3D Gaussian Splatting}
\author{Jiayu Ding}
\authornote{These authors contributed equally to this work.}
\affiliation{
	\institution{Peking University}
	\department{Shenzhen Graduate School}
	\department{Guangdong Provincial Key Laboratory of Ultra High Definition Immersive Media Technology}
	\city{Shenzhen}
	\country{China}
}
\email{jyding25@stu.pku.edu.cn}

\author{Meilu Song}
\authornotemark[1]
\affiliation{
    \institution{North China Electric Power University}
    \department{Department of Computer Science and Technology}
    \city{Baoding}
    \country{China}
}
\email{songml@ncepu.edu.cn}

\author{Yun Chen}
\authornotemark[1]
\affiliation{
    \institution{Hunan University}
    \department{College of Computer Science and Electronic Engineering}
    \city{Changsha}
    \country{China}
}
\email{cy2911@hnu.edu.cn}

\author{Wei Gao}
\affiliation{
	\institution{Peking University}
	\department{Shenzhen Graduate School}
	\department{Guangdong Provincial Key Laboratory of Ultra High Definition Immersive Media Technology}
	\city{Shenzhen}
	\country{China}
}
\email{gaowei262@pku.edu.cn}

\author{Ge Li}
\authornote{Corresponding author.}
\affiliation{
	\institution{Peking University}
	\department{Shenzhen Graduate School}
	\department{Guangdong Provincial Key Laboratory of Ultra High Definition Immersive Media Technology}
	\city{Shenzhen}
	\country{China}
}
\email{geli@ece.pku.edu.cn}

\renewcommand{\shortauthors}{Ding et al.}

\begin{document}

\begin{abstract}
While 3D Gaussian Splatting (3DGS) has advanced open vocabulary scene understanding, existing methods remain confined to explicit queries. They struggle to interpret implicit intents, complex spatial constraints, and commonsense reasoning required for practical embodied interactions. To address this gap, we introduce the task of reasoning 3D Gaussian segmentation and construct two benchmarks, Causal-LERF and Causal-ScanNet. These benchmarks systematically evaluate commonsense, spatial, affordance, and counterfactual reasoning. Evaluations reveal that current state of the art methods perform poorly on these reasoning challenges. Therefore, we propose CausalSplat, a framework that integrates vision-language models with 3D scene graphs to disentangle explicit structural perception from implicit logical inference. Extensive experiments demonstrate that CausalSplat achieves state of the art performance on our reasoning benchmarks while showing strong generalizability on standard referring and open vocabulary 3D segmentation tasks.
\textit{Project Page: https://jiayuding031020.github.io/CausalSplat}
\end{abstract}
\begin{teaserfigure}
	\centering
	\includegraphics[width=0.96\textwidth]{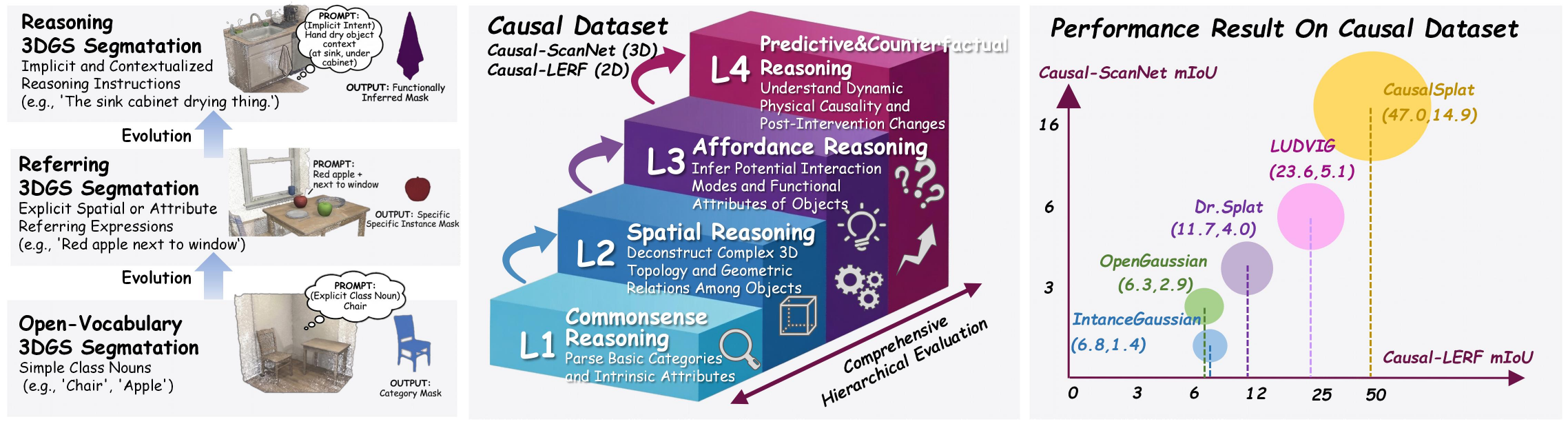}
	\caption{Overview of our reasoning segmentation paradigm and benchmark. Left: Evolution of 3DGS segmentation from basic open vocabulary queries and referring expressions to complex reasoning tasks driven by implicit intents. Middle: Our hierarchical framework evaluates models across four progressive levels: commonsense, spatial, affordance, and predictive/counterfactual reasoning. Right: CausalSplat significantly outperforms SOTA methods on 2D/3D metrics within the new causal dataset.}
	\label{fig:teaser}
\end{teaserfigure} 
\maketitle
\section{Introduction}
3D Gaussian Splatting~\cite{kerbl_3d_2023-1} (3DGS) is a leading neural rendering method known for its explicit representation and real-time rendering speed. Building on efficiency, recent studies advance open-vocabulary 3D scene understanding by distilling features from pretrained 2D vision-language models into 3D Gaussian fields. Despite this progress, current methods primarily handle explicit, noun-based queries. However, practical applications like Embodied AI~\cite{gupta_embodied_2021} require interpreting natural instructions with complex attributes. To address this, Referring 3D Gaussian Splatting~\cite{ReferSplat,ding2026zerosplat} (R3DGS) was proposed to localize objects using complex language descriptions involving spatial relations or specific attributes.

Current R3DGS methods rely on explicit referring expressions like category nouns or simple adjectives, struggling with complex reasoning. In real-world embodied interactions, user intents are highly implicit and context-dependent. Instead of rigid commands like ``find the blue towel,'' users often make situational requests, such as ``I just washed my hands, please hand me the thing to dry them next to the sink, hanging under the cabinet.'' Parsing these instructions requires inferring implicit affordances and decoding physical 3D spatial relations, demanding human-like commonsense and logical reasoning.
To address this, we introduce Reasoning 3D Gaussian Segmentation to accurately localize and segment objects in 3D scenes using complex, implicit language queries. Existing studies on implicit 3D understanding remain fragmented: 3DAffordSplat~\cite{wei20253daffordsplat} focuses solely on affordance in synthetic scenes, while REALM~\cite{shi2025realm} handles limited commonsense and spatial reasoning. We advocate for comprehensive abilities and propose a hierarchical taxonomy of 3D reasoning skills: (1) Commonsense Reasoning for basic categories and intrinsic attributes; (2) Spatial Reasoning for complex 3D geometric relations; (3) Affordance Reasoning for potential interactions and functions; and (4) Predictive and Counterfactual Reasoning for dynamic changes and causal relations under hypothetical conditions. Based on this framework, we build two benchmarks, Causal LERF and Causal ScanNet, derived from LERF~\cite{lerf2023} and ScanNet~\cite{dai2017scannet}. These datasets evaluate low-level perception and high-level reasoning by measuring fine-grained segmentation accuracy at 3D point and 2D pixel levels.

An intuitive solution for this new task is to follow the R3DGS approach, which involves assigning semantic features to individual 3D Gaussians and aligning them directly with text features. However, this direct method faces two main bottlenecks in reasoning tasks.
First, it lacks structured spatial awareness. Limited by the bag of words effect of models like CLIP~\cite{radford_learning_2021}, current feature alignment struggles to encode topological structures and relative geometric relations between objects. Thus, it fails to parse instructions with complex spatial constraints like orientation and inclusion.
Second, it lacks implicit reasoning. Traditional vision-language alignment focuses on explicit attributes and struggles to capture implicit causal logic and functional descriptions. This prevents building accurate logical mappings between abstract reasoning descriptions (e.g., tool to cut steak'') and concrete objects (e.g., knife'').

To tackle these challenges, we propose CausalSplat, a reasoning segmentation framework based on LLMs and scene graphs. CausalSplat conceptually separates the reasoning process into explicit structural perception and implicit logical inference. Specifically, we use scene graphs to explicitly model the topological and spatial relations among objects in 3D Gaussian scenes. This overcomes the spatial perception limitations of traditional feature fields. Meanwhile, we use the extensive knowledge and reasoning abilities of LLMs to parse implicit queries into executable structured intents. Through this design, CausalSplat effectively bridges the gap between low-level visual features and high-level logical semantics.

In summary, our main contributions are as follows: (1) We formally define a new task named Reasoning 3D Gaussian Segmentation. (2) We build two new reasoning oriented 3D scene datasets. These provide comprehensive benchmarks for evaluating spatial reasoning abilities under complex instructions. (3) We propose the CausalSplat framework, combining LLMs with scene graphs to effectively separate explicit structural perception from implicit logical inference. (4) Extensive experiments show that our method achieves state of the art performance on the proposed benchmarks.

\section{Related Works}

\subsection{3D Scene Representations}
Novel view synthesis and 3D scene reconstruction are fundamental tasks in computer vision. Neural Radiance Fields (NeRF) and its variants use implicit multi-layer perceptrons (MLPs) for high-quality novel view synthesis~\cite{mildenhall_nerf_2022}, significantly advancing 3D representation learning. However, the high computational cost of implicit volume rendering limits its application in real-time scenarios, such as embodied interaction. To address this, 3D Gaussian Splatting (3DGS) was proposed~\cite{kerbl_3d_2023-1}. 3DGS represents a 3D scene using explicit 3D Gaussian primitives. Each primitive is parameterized by a 3D center, a covariance matrix, view-dependent color, and opacity. For rendering, it uses an efficient differentiable rasterizer to project these primitives onto the 2D image plane, followed by depth-sorted alpha compositing to compute the final pixel colors. This explicit representation balances rendering quality and real-time performance while enabling direct discrete operations on scene geometry. Consequently, it provides a flexible structural foundation for the complex object-level logical reasoning tasks addressed in our work.

\subsection{Understanding in 3D Gaussian Splatting}
Enabling open-vocabulary understanding in 3D Gaussian fields has recently gained significant attention. Existing methods are primarily categorized into pixel-based and point-based approaches, both relying heavily on discriminative feature matching. Pixel-based methods~\cite{zhou_feature_2024-1,shi_language_2024,radford_learning_2021,zhang_dino_2022,qin_langsplat_2024,ye_gaussian_2024,qu_goi_2024} (e.g., LangSplat~\cite{qin_langsplat_2024} and Feature-3DGS~\cite{zhou_feature_2024-1}) distill features from 2D vision foundation models into scene-specific latent feature fields. They perform semantic matching at the pixel level in the rendered 2D image space, which often suffers from multi-view inconsistency. Conversely, point-based methods~\cite{wu2024opengaussian,li_instancegaussian_2025,lu_scaffold-gs_2024,jun-seong_dr_2025,marrie2025ludvig,ding2026extrinsplat} (e.g., OpenGaussian~\cite{wu2024opengaussian} and Dr.Splat~\cite{jun-seong_dr_2025}) lift 2D segmentation masks or CLIP features~\cite{radford_learning_2021} directly to 3D Gaussian centroids. This discretizes the scene into instance-aware clusters for point-level semantic matching.

Building on these foundations, recent studies explore deeper 3DGS understanding using natural language instructions. For instance, ReferSplat introduces referring 3D Gaussian understanding to localize targets using complex text descriptions~\cite{ReferSplat}. ZeroSplat further expands upon this task~\cite{ding2026zerosplat}. 3DAffordSplat focuses on fine-grained 3D affordance understanding by aligning cross-modal structures from point clouds to 3D Gaussians~\cite{wei20253daffordsplat}. REALM employs Multimodal Large Language Models (MLLMs) within an agent framework to perform open-world commonsense and implicit instruction reasoning on 3DGS~\cite{shi2025realm}. However, these pioneering works are typically tailored for single-dimensional reasoning and lack a systematic formulation of complex reasoning capabilities in 3D scenes. In contrast, our work introduces a multi-level reasoning framework encompassing commonsense, spatial, affordance, and predictive and counterfactual reasoning. This aims to comprehensively evaluate and systematically enhance the deep logical inference capabilities of models in continuous 3D scenes.

\begin{table*}[t]
	\centering
	\caption{Comparison of 3DGS-based understanding and reasoning datasets. Abbreviations used in the table: "Seg." for Segmentation, "Ref." for Referring, and "Inst." for Instructions. For reasoning capabilities, "Spa. Reas.", "Com. Reas.", "Aff. Reas.", and "Cou. Reas." represent Spatial, Commonsense, Affordance, and Counterfactual Reasoning, respectively.}
	\label{tab:dataset_comparison}
	\resizebox{\textwidth}{!}{
		\begin{tabular}{lllccccccccc}
			\toprule
			\textbf{Dataset} & \textbf{Source} & \textbf{Task Type} & \textbf{Domain} & \textbf{Scene} & \textbf{Scenes} & \textbf{Query Scale} & \textbf{Referring} & \textbf{Spa. Reas.} & \textbf{Com. Reas.} & \textbf{Aff. Reas.} & \textbf{Cou. Reas.} \\
			\midrule
			LERF & LangSplat~\cite{qin_langsplat_2024} & Open-Vocab Seg. & 2D & Real & 4 & 208 Words & - & - & - & - & - \\
			ScanNet & OpenGaussian~\cite{wu2024opengaussian} & Open-Vocab Seg. & 3D & Real & 10 & 84 Words & - & - & - & - & - \\
			Ref-LERF & ReferSplat~\cite{ReferSplat} & Ref. Seg. & 2D & Real & 4 & 63 Inst. & \checkmark & \checkmark & - & - & - \\
			3DAffordSplat & 3DAffordSplat~\cite{wei20253daffordsplat} & Reasoning Seg. & 3D & Synthetic & 8.3K & 6.6K Inst. & - & - & - & \checkmark & - \\
			REALM3D & REALM~\cite{shi2025realm} & Reasoning Seg. & 2D & Real & 100+ & 1K+ Pairs & \checkmark & \checkmark & \checkmark & - & - \\
			\rowcolor{gray!15} \textbf{Causal-LERF (Ours)} & \textbf{CausalSplat} & \textbf{Reasoning Seg.} & \textbf{2D} & \textbf{Real} & \textbf{4} & \textbf{158 Inst.} & \checkmark & \checkmark & \checkmark & \checkmark & \checkmark \\
			\rowcolor{gray!15} \textbf{Causal-ScanNet (Ours)} & \textbf{CausalSplat} & \textbf{Reasoning Seg.} & \textbf{3D} & \textbf{Real} & \textbf{10} & \textbf{73 Inst.} & \checkmark & \checkmark & \checkmark & \checkmark & \checkmark \\
			\bottomrule
		\end{tabular}
	}
\end{table*}
\subsection{Reasoning Segmentation}
Reasoning segmentation extracts accurate object masks from implicit text queries. LISA~\cite{lai2024lisa} introduced this by integrating large vision-language models with SAM~\cite{kirillov_segment_2023}. Later methods expanded this paradigm: PixelLM~\cite{ren2024pixellm} achieves pixel-level reasoning via a lightweight decoder and codebook; LLM-Seg~\cite{wang2024llm} bridges LLMs and SAM to filter mask proposals; LLaVASeg~\cite{yang2024empowering} applies chain-of-thought prompting for accurate segmentation and dialogue; and VISA~\cite{yan2024visa} extends implicit query-based tracking and segmentation to videos.
In the 3D domain, scene understanding has recently evolved to encompass a series of novel tasks~\cite{ding20263dinstructionambiguity,azuma2022scanqa,chen2020scanrefer,ma2022sqa3d}. Specifically for reasoning segmentation, existing paradigms primarily rely on point cloud representations. For example, PARIS3D~\cite{kareem2024paris3d} and Reasoning3D~\cite{chen2024reasoning3d} focus on part-level segmentation and interpretation of isolated 3D objects, making them difficult to apply to complex real-world 3D scenes. Recent methods like SegPoint~\cite{he2024segpoint} and Reason3D~\cite{huang2025reason3d} introduce LLM reasoning capabilities into 3D point cloud segmentation.
While recent studies have explored 3DGS as an alternative representation, their reasoning scope remains limited. For instance, 3DAffordSplat only investigates affordance reasoning for potential interaction regions in synthetic scenes~\cite{wei20253daffordsplat}, while REALM only addresses commonsense and limited spatial reasoning~\cite{shi2025realm}. In contrast, our method achieves systematic multi-level 3D Gaussian reasoning segmentation, providing an effective solution for complex embodied interaction in real-world scenes.
\begin{figure*}[t]
  \centering
  \includegraphics[width=\textwidth]{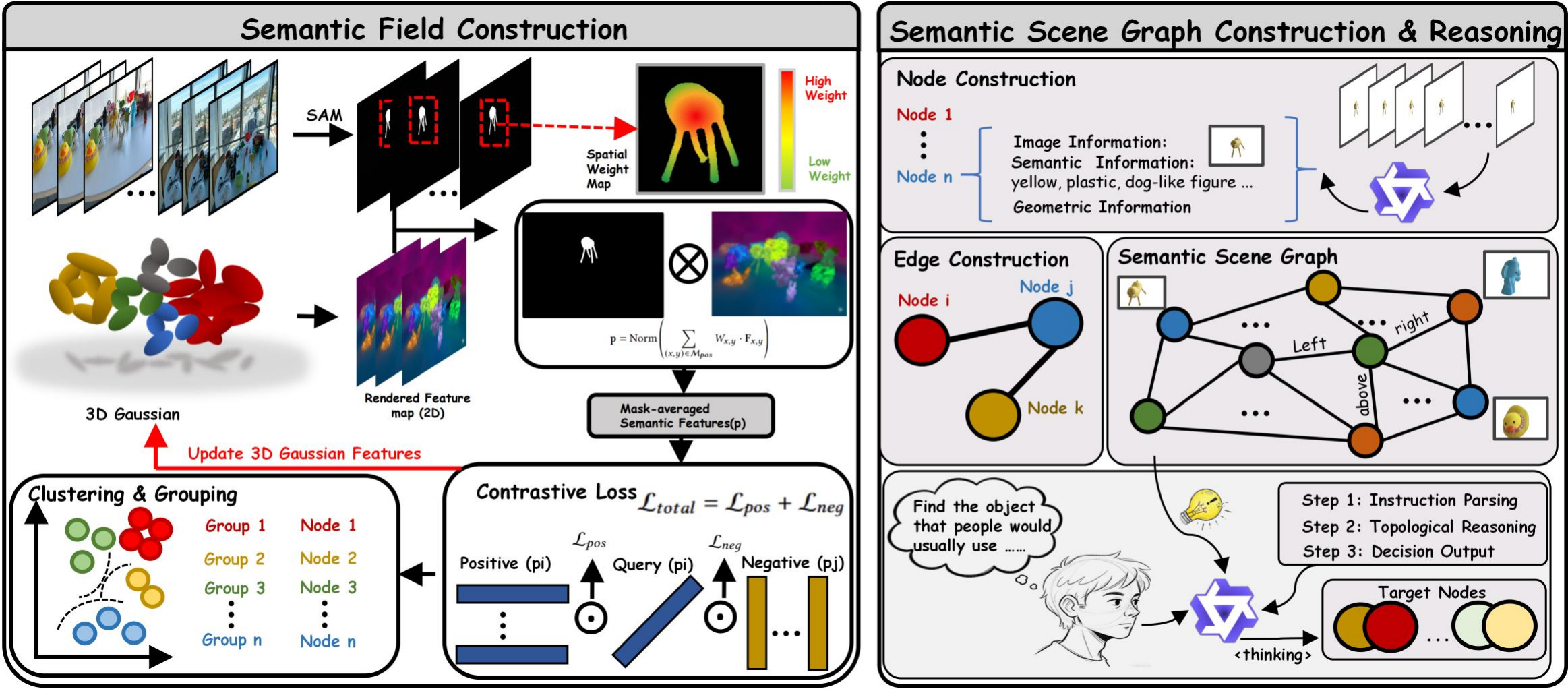}
  \caption{Overview of the proposed CausalSplat pipeline. (a) Semantic Field Construction. 2D masks are extracted via SAM, and mask-averaged semantic features $\mathbf{p}$ are obtained after spatial weighting and contrastive feature optimization. Afterward, 2D masks are clustered into 3D entities, and unique 3D entity labels are assigned to 3D Gaussian points. (b) Semantic Scene Graph Construction \& Reasoning. 3D entities are represented as nodes, interconnected via edges to form semantic scene graph. A VLM then performs a three-stage reasoning pipeline on thegraph to localize target nodes based on language instructions.}
  \label{fig:method_pipeline}
\end{figure*}

\section{Task and Benchmark}

\subsection{Task Formulation}
To advance 3D scene understanding toward higher-order logical reasoning, we formally define the task of \textbf{Reasoning 3D Gaussian Segmentation}. Given a 3D Gaussian scene and a natural language instruction with implicit intent or logic, the model must parse the instruction and accurately segment the target object in 3D space. Unlike traditional explicit referring expression segmentation, this task requires multi-level cognitive abilities, ranging from basic spatial perception to physical causal inference.

Although prior research has explored 3D reasoning, most works are confined to single-dimensional tasks, such as commonsense or affordance reasoning alone. Furthermore, the field lacks a benchmark for evaluating open-vocabulary 3D Gaussian understanding at the point level in real-world scenes.
To address these limitations, we introduce a multi-dimensional reasoning taxonomy and construct two new benchmark datasets: \textbf{Causal-LERF} for 2D pixel-level evaluation and \textbf{Causal-ScanNet} for 3D point-level evaluation. This is the first multi-level reasoning benchmark specifically built for 3DGS, covering dimensions from basic spatial topology to physical counterfactual inference.

\subsection{The Hierarchical Reasoning Taxonomy}
\label{sec:taxonomy}
Existing 3D understanding datasets often focus on isolated dimensions, failing to provide a systematic evaluation of comprehensive reasoning. We categorize real-world interaction instructions into four progressive levels:
\noindent \textbf{1) Spatial Reasoning:} Requires parsing absolute and relative geometric relations (e.g., distance, scale) and identifying 3D topologies like support, containment, and occlusion. For example, ``Find the object occluded under the table that supports a box.'' This requires 3D structural perception to overcome 2D limitations.
\noindent \textbf{2) Commonsense Reasoning:} Requires aligning LLM prior knowledge with 3D scene instances to interpret implicit semantics without direct visual descriptions. For example, given ``I just finished a greasy meal and need to clean my mouth,'' the model must use commonsense to locate a napkin based on functional relevance rather than appearance matching.
\noindent \textbf{3) Affordance Reasoning:} Requires inferring whether an object supports embodied actions based on its 3D geometry (e.g., concavity, handle width) and spatial state. For example, ``Find a container with an upward opening for a robotic arm to grasp.'' This demands precise mapping between physical structure and utility.
\noindent \textbf{4) Predictive and Counterfactual Reasoning:} As the highest cognitive level, this requires understanding mechanisms governed by physics (e.g., gravity, collision) to perform counterfactual deduction. For example, ``If the bottom red object is removed, which objects will fall?'' The model must predict state changes by analyzing 3D force distribution and structural dependencies.

\subsection{Dataset}
\label{sec:dataset}
\noindent\textbf{Annotation and Quality Control}
To ensure accuracy and logical rigor, all instructions undergo strict generation and verification:
\textit{1) Instruction Initialization:} We use ground truth object annotations from real 3D scans and preset templates to guide an LLM in generating initial instructions across all four levels.
\textit{2) Physical and Logical Consistency Check:} Four professional annotators review the instructions from multiple perspectives. An instruction is included only if all four annotators approve it; ambiguous samples are removed.

\noindent\textbf{Dataset Statistics and Splits}
Causal-LERF and Causal-ScanNet contain a total of 231 reasoning instructions across 14 real 3D indoor scenes. The instruction distribution consists of 18.2\% spatial reasoning, 48.5\% commonsense reasoning, 16.0\% affordance reasoning, and 17.3\% predictive and counterfactual reasoning.

\subsection{Evaluation Metrics}
\label{sec:metrics}
We define Reasoning 3D Gaussian Segmentation as a target localization and mask segmentation task. Accordingly, we adopt two primary metrics: 2D Mean Intersection over Union (mIoU) for pixel-level evaluation on Causal-LERF, and 3D mIoU for point-level evaluation on Causal-ScanNet.

\section{Method}
Given multi-view 2D observations and a natural language instruction, we aim to localize a target object specified by complex intents within a 3D Gaussian scene. As illustrated in Figure~\ref{fig:method_pipeline}, our framework consists of three modules: (1) \textbf{Semantic Field Construction}, which lifts 2D segmentation masks into a 3D feature field; (2) \textbf{Multimodal Semantic Scene Graph Construction}, which structures 3D entities into a scene graph with multimodal attribute nodes and scale adaptive topological edges; and (3) \textbf{Scene Graph based Multimodal Reasoning}, which guides the model to parse instructions and perform a topological search within the graph to output the target entity.

\subsection{Semantic Field Construction}
This module lifts multi-view 2D segmentation masks into a 3D semantic feature field. To mitigate inherent edge noise in 2D segmentation masks, our method proceeds in three stages: suppressing edge noise via \textbf{spatially weighted feature extraction} to obtain mask-averaged semantic features; aggregating target features and separating backgrounds in the 3D feature space via \textbf{contrastive feature optimization}; and performing clustering based \textbf{3D instance assignment} to group discrete 2D segmentation masks into unified 3D entities. This completes the instance assignment for 3D Gaussian points.

\noindent \textbf{Spatially Weighted Feature Extraction.} To represent semantic information in 3D space, we associate each 3D Gaussian point with a semantic feature vector $\mathbf{f}_i \in \mathbb{R}^C$. Following the differentiable rendering pipeline of 3DGS [10], the rendered feature map $\mathbf{F} \in \mathbb{R}^{H \times W \times C}$ is generated by splatting these 3D features onto the 2D image plane using the same alpha-blending process as the color rendering. Specifically, the rendered feature $\mathbf{F}_{x,y}$ at pixel $(x,y)$ is formulated as the weighted aggregation of semantic feature vector $\mathbf{f}_i$ from the ordered set of overlapping Gaussians $\mathcal{N}$:
\begin{equation}
\mathbf{F}_{x,y} = \sum_{i \in \mathcal{N}} \mathbf{f}_i \alpha_i \prod_{j=1}^{i-1} (1 - \alpha_j)
\end{equation}
where $\mathcal{N}$ denotes the sorted Gaussians overlapping pixel $(x,y)$ according to their depth. Simultaneously, given multi-view images, we utilize the SAM model [13] to extract a set of 2D segmentation masks $\mathcal{M}_v$ for each view $v \in \mathcal{V}_{train}$, providing the input for subsequent instance assignment.
However, the mask boundaries generated by SAM often contain inherent semantic noise, which can degrade the quality of the reconstructed 3D feature field. To extract 3D feature representations from these noisy observations, we propose a spatial weighting strategy. Given the rendered feature map $\mathbf{F}$ and its corresponding instance mask $M_{pos} \in \mathcal{M}_v$, we first compute geometric center $(c_x, c_y)$ of the mask. The spatial weight $W_{x,y}$ for each pixel $(x,y)$ within the mask $M_{pos}$ is then defined as:
\begin{equation}
W_{x,y} = \omega_{min} + (1 - \omega_{min}) \left(1 - \frac{d_{x,y}}{d_{max} + \epsilon}\right)
\end{equation}
where $d_{x,y}$ is the Euclidean distance from pixel $(x,y)$ to the geometric center, $d_{max}$ is the maximum distance from any pixel in $M_{pos}$ to the geometric center, and $\omega_{min}$ is a predefined minimum weight threshold. Using this weight, we aggregate the features within $M_{pos}$ to extract a mask-averaged semantic feature $\mathbf{p} \in \mathbb{R}^{C}$ representing the instance:
\begin{equation}
\mathbf{p} = \text{Norm} \left( \sum_{(x,y) \in \mathcal{M}_{pos}} W_{x,y} \cdot \mathbf{F}_{x,y} \right)
\end{equation}
where $\text{Norm}(\cdot)$ denotes $L_2$ normalization. We then compute the cosine similarity map $\mathbf{S}$ between the mask-averaged semantic features $\mathbf{p}$ and the global feature field to guide the subsequent stages of the contrastive feature optimization.

\noindent \textbf{Contrastive Feature Optimization.} To ensure semantic consistency, we design an end-to-end contrastive learning objective~\cite{khosla2020supervised} $\mathcal{L}_{total}$, consisting of a positive alignment term $\mathcal{L}_{pos}$ and a negative repulsion term $\mathcal{L}_{neg}$:
\begin{equation}
\mathcal{L}_{total} = \mathcal{L}_{pos} + \mathcal{L}_{neg}
\end{equation}
The positive alignment term applies the spatial weight $W_{x,y}$ to attenuate boundary noise, guiding the target region features to cluster toward the mask-averaged semantic features:
\begin{equation}
\mathcal{L}_{pos} = \frac{\sum_{(x,y) \in \mathcal{M}_{pos}} W_{x,y} (1 - S_{x,y})^2}{\sum_{(x,y) \in \mathcal{M}_{pos}} W_{x,y}}
\end{equation}
The negative repulsion term constrains  the negative sample features using a margin $m$. It penalizes the negative samples with a similarity exceeding $m$ to increase the discriminability between the positive sample and the negative sample:
\begin{equation}
\mathcal{L}_{neg} = \frac{1}{|\mathcal{M}_{neg}|} \sum_{(x,y) \in \mathcal{M}_{neg}} \ell_{neg}(S_{x,y})
\end{equation}
where the computed penalty term is $\ell_{neg}(S_{x,y}) = (S_{x,y} - m)^2$ if $S_{x,y} > m$, and $0$ otherwise.

Because negative sample regions dominate the image, using all negative samples causes severe class imbalance. We therefore introduce a dynamic sampling strategy. An indicator function $\mathbb{I}_{j}$ filters the negative sample set $\mathcal{M}_{neg}$ during optimization:
\begin{equation}
\mathbb{I}_{j} =
\begin{cases}
	1, & \text{if } \eta_j < r \text{ or } S_j > \tau \\
	0, & \text{otherwise}
\end{cases}
\end{equation}
where $\eta_j \sim \mathcal{U}(0, 1)$ is a uniformly distributed sampling factor, and $\tau$ is the hard negative threshold. The sampling ratio $r = N_{pos} / N_{neg}$ is the pixel count ratio of positive to negative samples, where $N_{pos} = \sum_{i \in \Omega} \mathbb{I}(i \in \mathcal{M}_{pos})$ and $N_{neg} = \sum_{j \in \Omega} \mathbb{I}(j \in \mathcal{M}_{neg})$.
\begin{figure*}[t]
	\centering
	\includegraphics[width=\textwidth]{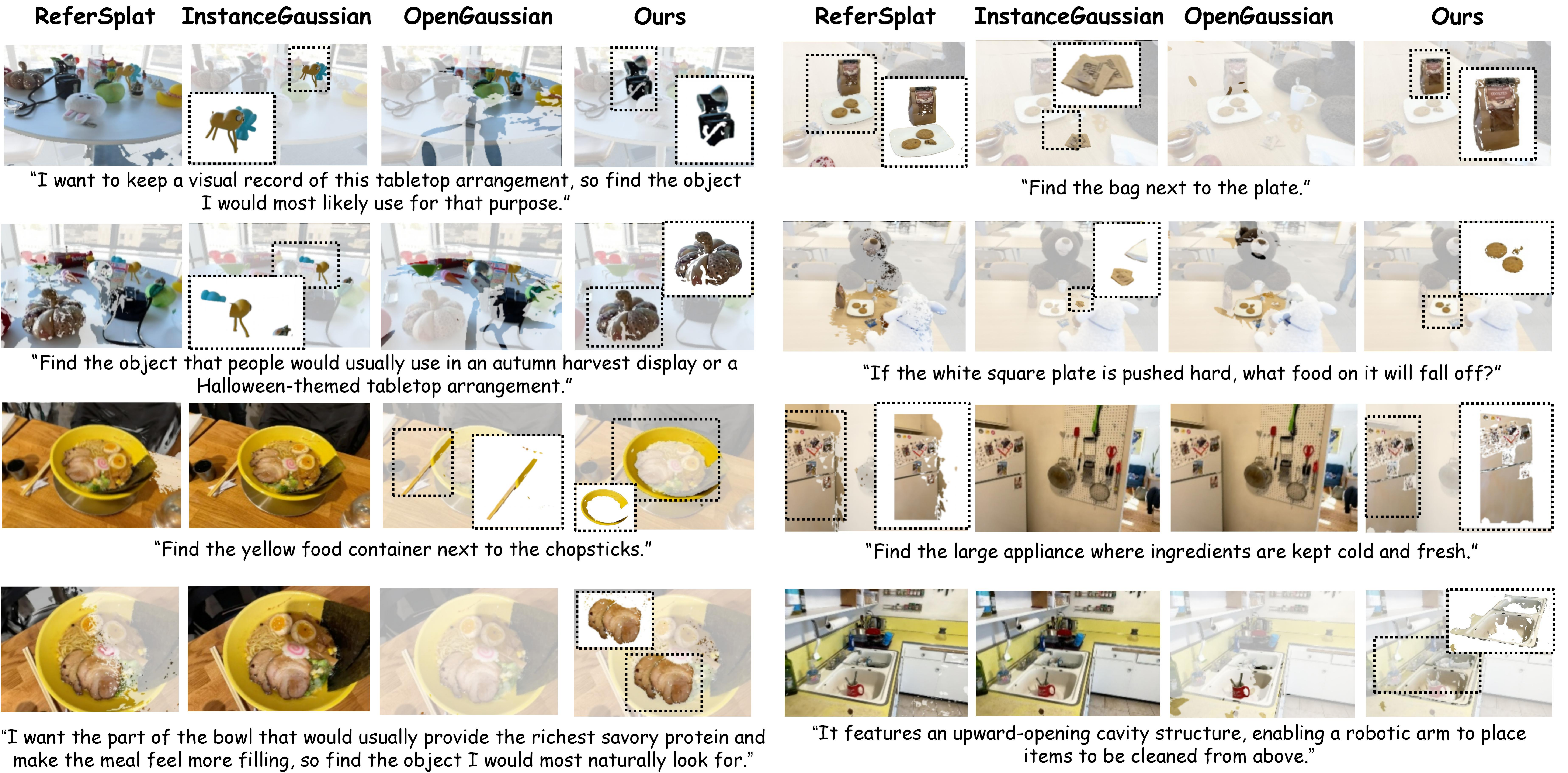}
	\caption{Qualitative comparison on the Causal-LERF dataset.}
	\label{fig:q1}
\end{figure*}

\noindent \textbf{3D Instance Assignment.} To link 2D segmentation masks to 3D entities, we employ a feature clustering strategy. Following the contrastive feature optimization, each 2D mask is associated with a mask-averaged semantic feature. We collect mask-averaged semantic features from all views and apply HDBSCAN algorithm for density based clustering. This groups 2D segmentation masks into 3D entity clusters $\mathcal{O}_k$.

Next, we compute representative feature $\mathbf{C}_k$ for each 3D entity as the mean of all mask-averaged semantic
features in the cluster:
\begin{equation}
\mathbf{C}_k = \text{Norm}\left( \frac{1}{|\mathcal{O}_k|} \sum_{\mathbf{p} \in \mathcal{O}_k} \mathbf{p} \right)
\end{equation}
Finally, we use $\mathbf{C}_k$ to assign 3D instances to the 3D Gaussian point cloud $\mathcal{P}$. For any 3D Gaussian point $j \in \mathcal{P}$, we compute the cosine similarity between its semantic feature vector $\mathbf{f}_j$ and representative features of all 3D entities to obtain the score $\mathbf{e}_{j,k} = \mathbf{f}_j \cdot \mathbf{C}_k^\top$. Following a hard assignment principle, the $j$-th Gaussian point $j \in \mathcal{P}$ is uniquely assigned to a specific 3D entity ID by selecting the highest similarity score. Consequently, the $i$-th 3D physical entity is represented by the point set $\mathcal{P}_i = \{j \in \mathcal{P} \mid \arg\max_k (e_{j,k}) = i\}$

\subsection{Semantic Scene Graph Construction}
To enable Vision-Language Models (VLMs) to parse 3D scenes, we transform extracted 3D entities into structured multimodal semantic scene graph $\mathcal{G} = (\mathcal{V}, \mathcal{E})$. Nodes $\mathcal{V}$ represent 3D entities, and edges $\mathcal{E}$ represent spatial relationships between them.

\noindent \textbf{Semantic Node Construction.} We instantiate each extracted 3D entity as a graph node $n_i \in \mathcal{V}$. To align 3D physical space with textual semantics, we build a multimodal descriptor $\mathcal{M}_i = \langle \mathbf{c}_i, \mathbf{b}_i, \mathcal{I}_i, \mathcal{A}_i \rangle$ for each node. Geometrically, the spatial centroid $\mathbf{c}_i \in \mathbb{R}^3$ is the mean coordinate of all Gaussian points in the 3D entity. The axis-aligned bounding box span $\mathbf{b}_i \in \mathbb{R}^3$ is the coordinate range of the point cluster $\mathcal{P}_i \subseteq \mathcal{P}$ across each dimension: $\mathbf{b}_i = \max(\mathcal{P}_i) - \min(\mathcal{P}_i)$, where $\mathcal{P}_i$ denotes the set of Gaussian points assigned to the $i$-th entity. $\mathcal{I}_i$ represents 2D segmentation masks associated with the 3D entity. Semantically, we use a Vision-Language Model (VLM) to analyze 2D segmentation masks of the 3D entity and extract a structured tuple $\mathcal{A}_i = \{ \mathcal{T}_i, \mathcal{X}_i, \mathcal{F}_i \}$. This tuple records the intrinsic category, visual features, and interactive affordances of the 3D entity.

\noindent \textbf{Scale Adaptive Edge Construction.} Open-world scenes often exhibit drastic scale variations, making fixed distance thresholds unreliable for determining spatial relationships. We propose an adaptive topological edge construction mechanism. Using vertical relationships as an example, we require 3D entities to have significant displacement along the vertical axis while strictly limiting horizontal deviation. First, the relative displacement $\Delta Z = z_i - z_j$ between the centroids of two entities $\{\mathcal{P}_i, \mathcal{P}_j\}$ (represented by nodes $n_i, n_j$) along the $Z$-axis must be a large proportion of their centroid distance $d_{i,j}$, satisfying $|\Delta Z| / d_{i,j} > \tau$. Second, their projected distance on the horizontal $XY$-plane must be smaller than a dynamic tolerance $\mathcal{D}_{limit}^{xy}$:
\begin{equation}
\mathcal{D}_{limit}^{xy} = \min \left( \max \left( 0.55 \sqrt{\bar{b}_{x}^2 + \bar{b}_{y}^2}, 0.08 \mathcal{S}_{xy} \right), 0.35 \mathcal{S}_{xy} \right)
\end{equation}
where $\bar{b}_x$ and $\bar{b}_y$ are the average spans of the two entities' 3D bounding boxes along the corresponding axes:
\begin{equation}
\begin{aligned}
\bar{b}_x &= \frac{b_{i,x} + b_{j,x}}{2}, & \quad \bar{b}_y &= \frac{b_{i,y} + b_{j,y}}{2}
\end{aligned}
\end{equation}
and $\mathcal{S}_{xy} = \sqrt{(\Delta X_{scene})^2 + (\Delta Y_{scene})^2}$ represents the diagonal span of the entire 3D scene on the horizontal plane. This scene span is calculated from the extrema of the centroid coordinate set $\mathcal{C}$ for all  3D entity valid nodes:
\begin{equation}
\begin{aligned}
\Delta X_{scene} &= \max(\mathcal{C}_x) - \min(\mathcal{C}_x), & \quad \Delta Y_{scene} &= \max(\mathcal{C}_y) - \min(\mathcal{C}_y)
\end{aligned}
\end{equation}
The dynamic tolerance balances object size and overall scene range.

\begin{table}[t]
  \centering
  \caption{Quantitative comparison on the Causal-LERF dataset. We report the 2D mIoU (\%) metric.}
  \label{tab:causal_lerf}
  \resizebox{\linewidth}{!}{
  \begin{tabular}{lcccccc}
    \toprule
    Method & Ramen & Teatime & Figurines & Waldo & Mean \\
    \midrule
    OpenGaussian~\cite{wu2024opengaussian} & 9.6 & 4.2 & 3.3 & 8.2 & 6.3 \\
    Dr.Splat~\cite{jun-seong_dr_2025} & 8.7 & 8.8 & 16.1 & \underline{13.3} & 11.7 \\
    InstanceGaussian~\cite{li_instancegaussian_2025} & 2.7 & 4.7 & 9.2 & 10.5 & 6.8 \\
    LUDVIG~\cite{marrie2025ludvig} & \textbf{32.6} & \underline{19.7} & \underline{34.1} & 8.0 & \underline{23.6} \\
    ReferSplat~\cite{ReferSplat} & 5.2 & 16.2 & 7.5 & 11.1 & 10.0 \\
    REALM~\cite{shi2025realm} & 9.4 & 14.3 & 7.8 & 13.9 & 11.4 \\
    \rowcolor{gray!20} Ours & \underline{26.2} & \textbf{68.4} & \textbf{46.9} & \textbf{46.5} & \textbf{47.0} \\
    \bottomrule
  \end{tabular}
  }
\end{table}
\subsection{Multimodal Reasoning}
 After constructing the structured scene graph, we use a Vision-Language Model with prompt engineering to precisely localize the target 3D entity specified by the natural language instruction $Q$. Specifically, We utilize the semantic scene graph $\mathcal{G}$ as the multimodal input. and guide the model through a three-stage reasoning pipeline: \textbf{1) Instruction Parsing:} The model first parses the semantics of instruction $Q$. If the instruction only describes the intrinsic features of an object, the model skips to decision output. If explicit spatial relationship words are detected, topological reasoning is triggered.
 \textbf{2) Topological Reasoning:} The model identifies the anchor node $v_{anchor}$ described in the instruction within the scene graph $\mathcal{G}$, and performs a structured search along the directed edges $\mathcal{E}$.
 \textbf{3) Decision Output:} The model integrates semantic and topological constraints to output the final target node. Detailed prompt designs are provided in the supplementary material.

\begin{figure*}[t]
	\centering
	\includegraphics[width=\textwidth]{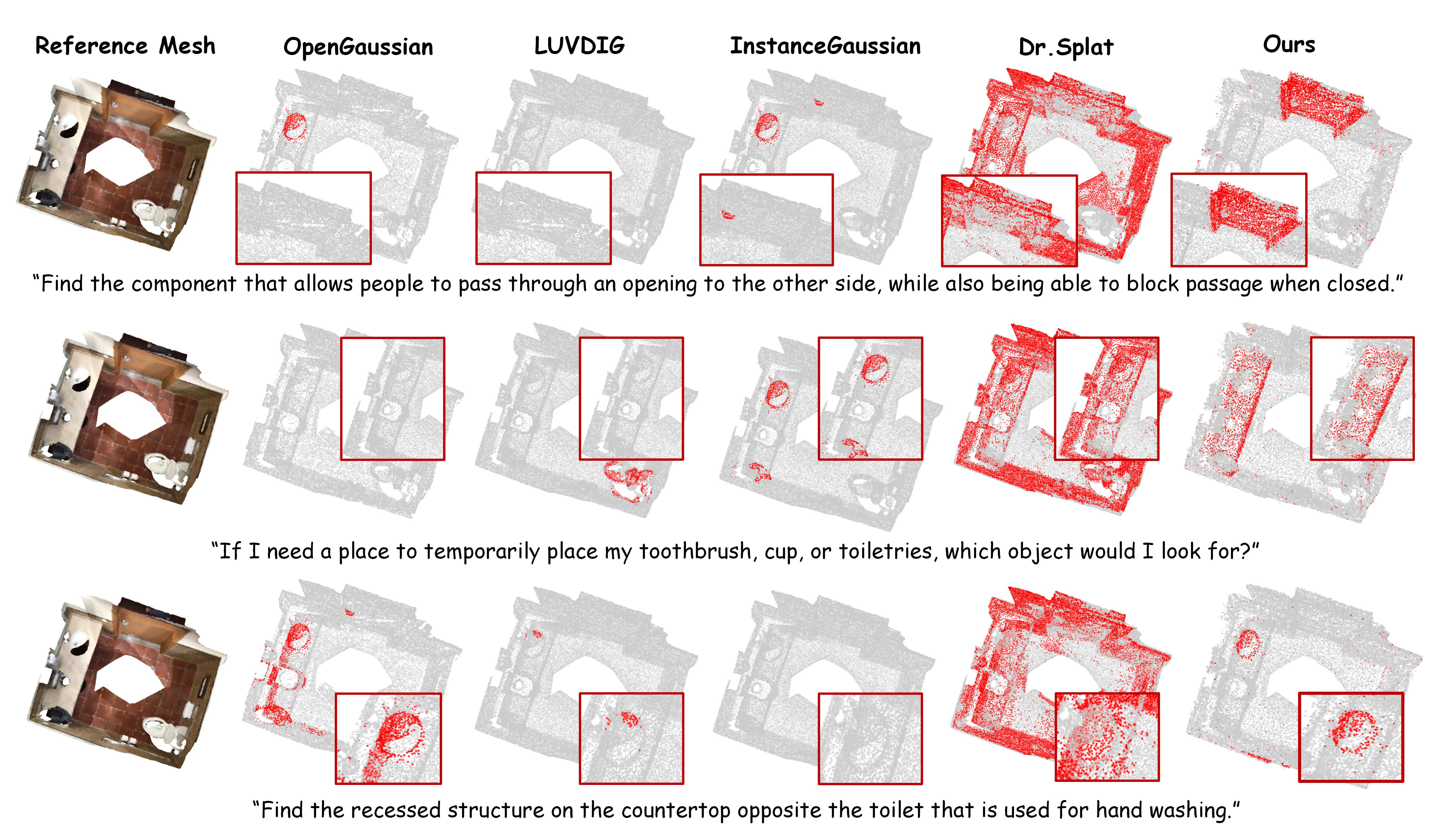}
	\caption{Qualitative comparison on the Causal-ScanNet dataset.}
	\label{fig:q2}
\end{figure*}

\begin{table}[t]
	\centering
	\caption{Quantitative comparison on the Causal-ScanNet dataset. We report the 3D mIoU (\%) metric.}
	\label{tab:causal_scannet}
	\begin{tabular}{p{5cm}c}
		\toprule
		Method & mIoU \\
		\midrule
		OpenGaussian~\cite{wu2024opengaussian} & 2.9 \\
		InstanceGaussian~\cite{li_instancegaussian_2025} & 1.4 \\
		Dr.Splat~\cite{jun-seong_dr_2025} & 4.0 \\
		LUDVIG~\cite{marrie2025ludvig} & \underline{5.1} \\
		\rowcolor{gray!20} Ours & \textbf{14.9} \\
		\bottomrule
	\end{tabular}
\end{table}

\section{Experiments}
\subsection{Implementation Details}
We adopt SAM for 2D mask extraction and HDBSCAN for 3D instance clustering. Key hyperparameters: \(\omega_{min}=0.4\), contrast margin \(m=0.03\), hard negative threshold \(\tau=0.5\), vertical edge threshold 0.3. All semantic features are L2-normalized. We use Qwen3-VL-30B-A3B-Instruct to build scene graph attributes and parse text queries.

\subsection{Reasoning 3D Gaussian Segmentation}

\noindent\textbf{Settings.} \textbf{1) Task.} This task requires models to localize and segment a target within a 3D scene based on natural language queries containing implicit intents or multi-hop logical constraints. Consequently, it demands capabilities beyond basic object recognition, requiring robust common-sense reasoning and spatial topological analysis. \textbf{2) Baselines.} We compare our approach against three categories of 3D scene understanding methods: open-vocabulary understanding, referring understanding, and reasoning understanding. \textbf{3) Metrics.} We evaluate segmentation accuracy using the mean Intersection over Union (mIoU). Specifically, we report the 2D mIoU for the Causal-LERF dataset and the 3D mIoU for the Causal-ScanNet dataset to properly measure the spatial overlap between the predictions and the ground-truth annotations.

\noindent\textbf{Results on Causal-LERF}
As presented in Table~\ref{tab:causal_lerf}, our method achieves the highest performance on the Causal-LERF dataset with an average mIoU of 47.0\%. This significantly outperforms the second-best method, LUDVIG (23.6\% mIoU). In complex environments like \textit{Teatime} and \textit{Waldo}, our model surpasses baselines by a large margin. Specifically, open-vocabulary and referring methods lack the reasoning capabilities required to interpret implicit intents. Furthermore, reasoning methods like REALM struggle with highly complex logical constraints. These results emphasize the need for advanced spatial reasoning in accurate 3D segmentation.

Figure~\ref{fig:q1} provides qualitative comparisons on Causal-LERF. The visualizations show that our approach accurately segments targets described by complex queries with implicit intents and multi-hop spatial constraints. For queries requiring common-sense reasoning (e.g., identifying a camera for a ``visual record'' or a pumpkin for an ``autumn harvest display''), baseline models frequently misunderstand the semantics and incorrectly localize salient but irrelevant distractors. Additionally, under multi-hop spatial conditions (e.g., ``the bag next to the plate''), referring and open-vocabulary methods often fail to separate the primary target from the reference object. This leads to severe over-segmentation or incorrect localization. In contrast, our model effectively grounds abstract logical conditions and complex spatial relationships into the physical 3D space, consistently producing accurate boundaries.

\noindent\textbf{Results on Causal-ScanNet}
Table~\ref{tab:causal_scannet} details the quantitative evaluation on the more complex Causal-ScanNet dataset. Our method achieves an mIoU of 14.9\%, nearly tripling the performance of the second-best method, LUDVIG (5.1\%). The low performance of existing open-vocabulary models highlights the extreme difficulty of this task, which requires robust common-sense reasoning amid dense distractors in large-scale point clouds. This substantial gain confirms the robustness and scalability of our reasoning framework.

As illustrated in Figure~\ref{fig:q2}, our method effectively handles the complex spatial geometries and semantic ambiguities of the Causal-ScanNet dataset. When given queries about object affordances or implicit functions (e.g., an object allowing passage or a surface for toiletries), existing models struggle to ground these abstract concepts. This typically results in missed predictions, scattered noise, or severe over-segmentation across the room, as seen with Dr.Splat. Moreover, for queries with dense spatial constraints (e.g., locating a structure opposite the toilet), baselines frequently misidentify the reference object as the target. Conversely, our model successfully interprets these spatial and functional constraints to output precise segmentation masks directly on the 3D point clouds.

\subsection{Referring 3D Gaussian Segmentation}
\noindent\textbf{Settings.} \textbf{1) Task.} This task aims to precisely localize and segment a single target object in a 3D scene based on natural language instructions. \textbf{2) Baselines.} We primarily compare CausalSplat with ReferSplat, the current SOTA method for this task. For a broader evaluation, we also include relevant baselines. \textbf{3) Datasets.} We evaluate on the standard benchmark provided by ReferSplat.

\noindent\textbf{Results.} Table \ref{tab:ref_lerf_gr3dgs} shows our method achieves a state-of-the-art mean mIoU of 36.1 on Ref-LERF. This surpasses the previous best, ReferSplat, by 6.9 points. Although our method is not specifically designed for this task, it still performs favorably, demonstrating strong generalization ability.

\begin{table}[t]
	\centering
	\caption{Quantitative results for R3DGS on Ref-LERF measured by mIoU.}
	\label{tab:ref_lerf_gr3dgs}
	\resizebox{\linewidth}{!}{
		\begin{tabular}{l c c c c c}
			\toprule
			Method & Ramen & Teatime & Figurines & Waldo & Mean \\
			\midrule
			Grounded SAM~\cite{ren2024grounded} & 14.1 & 16.9 & 16.0 & 16.2 & 15.8 \\
			LangSplat~\cite{qin_langsplat_2024}    & 12.0 &  7.6 & 17.9 & 17.9 & 13.9 \\
			SPIn-NeRF~\cite{mirzaei2023spin}    &  7.3 & 11.7 &  9.7 & 10.3 &  9.8 \\
			GS-Grouping~\cite{ye_gaussian_2024}  & 27.9 & 14.8 &  8.6 &  6.3 & 14.4 \\
			GOI~\cite{qu_goi_2024}          & 27.1 & 22.9 & 16.5 & 15.7 & 20.5 \\
			ReferSplat~\cite{ReferSplat}    & \textbf{35.2} & \underline{31.3} & \underline{25.7} & \underline{24.4} & \underline{29.2} \\
			\rowcolor{gray!20} \textbf{Ours} & \underline{28.2} & \textbf{38.0} & \textbf{50.6} & \textbf{27.7} & \textbf{36.1} \\
			\bottomrule
		\end{tabular}
	}
\end{table}

\begin{table}[t]
	\centering
	\caption{Quantitative results for open-vocabulary object selection on LERF measured by mIoU.}
	\label{tab1}
	\resizebox{\linewidth}{!}{
		\begin{tabular}{l c c c c c}
			\toprule
			Method & Ramen & Teatime & Figurines & Waldo & Mean \\
			\midrule
			\textbf{Pixel-based} & & & & & \\
			LEGaussians~\cite{shi_language_2024} & 46.0 & 60.3 & 40.8 & 39.4 & 46.6 \\
			LangSplat~\cite{qin_langsplat_2024} & 51.2 & 65.1 & 44.7 & 44.5 & 51.4 \\
			Feature-3DGS~\cite{zhou_feature_2024-1} & 43.7 & 58.8 & 40.5 & 39.6 & 45.7 \\
			GS-Grouping~\cite{ye_gaussian_2024} & 45.5 & 60.9 & 40.0 & 38.7 & 46.3 \\
			GOI~\cite{qu_goi_2024} & 52.6 & 63.7 & 44.5 & 41.4 & 50.6 \\
			ReferSplat & \underline{55.1} & 50.1 & \textbf{67.5} & 48.9 & 55.4 \\
			Occam's LGS~\cite{Cheng_2025_BMVC} & 51.0 & \underline{70.2} & \underline{58.6} & \textbf{65.3} & \underline{61.3} \\
			3DVLGS~\cite{peng20243d} & \textbf{61.4} & \textbf{73.5} & 58.1 & \underline{54.8} & \textbf{62.0} \\
			\midrule
			\textbf{Point-based} & & & & & \\
			OpenGaussian~\cite{wu2024opengaussian} & \underline{31.0} & 60.4 & 39.3 & 22.7 & 38.4 \\
			InstanceGaussian~\cite{li_instancegaussian_2025} & 24.6 & \underline{63.4} & 45.5 & 29.2 & 40.7 \\
			Dr.Splat(Top-40)~\cite{jun-seong_dr_2025} & 24.7 & 57.2 & 53.4 & \underline{39.1} & 43.6 \\
			LUDVIG~\cite{marrie2025ludvig} & \textbf{42.3} & 58.6 & \underline{58.0} & \textbf{42.8} & \underline{50.4} \\
			\rowcolor{gray!20} \textbf{Ours} & 26.7 & \textbf{73.0} & \textbf{75.3} & 30.1 & \textbf{51.3} \\
			\bottomrule
		\end{tabular}
	}
\end{table}

\subsection{Open-Vocabulary 3D Gaussian Segmentation}

\noindent\textbf{Settings.} \textbf{1) Task:} This task aims to precisely localize and segment a single target object within a 3D scene based on a given text category label.
\textbf{2) Datasets:} We conduct experiments on the LERF-OVS dataset re-annotated by LangSplat.
\noindent\textbf{3) Baselines:} We compare our approach against twelve open-vocabulary 3D baselines, categorized into pixel-based and point-based methods.

\noindent\textbf{Results.} As shown in Table~\ref{tab1}, CausalSplat achieves a state-of-the-art mIoU of 51.3 among point-based methods on LERF. These results demonstrate that our semantic field construction enables robust feature extraction and precise instance decoupling, proving effective for standard open-vocabulary segmentation despite being designed for complex reasoning.

\subsection{Ablation Studies}
We conduct comprehensive ablation studies to evaluate the key components of our framework. All experiments use the mIoU metric, and the quantitative results are summarized in Table \ref{tab:ablation_merged}.

\noindent \textbf{Semantic Field Construction.} Standard average pooling introduces 2D boundary noise, yielding 35.7 mIoU. Our spatial weighting suppresses this noise and preserves 3D consistency, boosting mIoU to 47.0. Random weighting performs worst (34.9), confirming that structured feature aggregation is essential.

\noindent \textbf{Scene Graph Construction.} A text-only graph achieves 40.1 mIoU. Adding only images degrades performance (37.2), while adding only edges provides minimal gains (40.2). Combining both visual nodes and spatial edges achieves 47.0 mIoU, proving that these modalities require joint modeling for scene understanding.

\noindent \textbf{VLM Prompt Design.} Skipping instruction parsing drops mIoU to 40.5, as redundant searches cause errors on simple queries. Omitting topological reasoning drops mIoU to 42.2 due to failures on complex spatial tasks. The complete pipeline dynamically routes reasoning to reach 47.0 mIoU.

\begin{table}[t]
	\centering
	\caption{Ablation studies on key framework components measured by mIoU.}
	\label{tab:ablation_merged}
	\resizebox{0.85\linewidth}{!}{
		\begin{tabular}{p{6cm} c}
			\toprule
			\textbf{Component Configuration} & \textbf{mIoU} \\
			\midrule
			\textbf{Semantic Field Construction} & \\
            Random Weighting  & 34.9 \\
			Average Pooling  & 35.7 \\
			Spatial Weighting  & \textbf{47.0} \\
			\midrule
			\textbf{Scene Graph Construction} & \\
			Text Nodes Only & 40.1 \\
			Text Nodes + Node Images & 37.2 \\
			Text Nodes + Topological Edges & 40.2 \\
			Full Multimodal Graph & \textbf{47.0} \\
			\midrule
			\textbf{VLM Reasoning Prompt} & \\
			Without Instruction Parsing & 40.5 \\
			Without Topological Reasoning & 42.2 \\
			Complete CoT Pipeline & \textbf{47.0} \\
			\bottomrule
		\end{tabular}
	}
\end{table}

\section{Conclusion}
In this paper, we introduce the novel task of Reasoning 3D Gaussian Segmentation to bridge the gap between basic 3D perception and higher-order logical reasoning. To systematically evaluate this, we present two comprehensive benchmarks: Causal-LERF and Causal-ScanNet. Furthermore, we propose CausalSplat, a framework that synergizes Vision-Language Model (VLM) with 3D semantic scene graphs to disentangle explicit structural perception from implicit logical inference. Extensive experiments demonstrate that CausalSplat not only establishes a new state-of-the-art on complex reasoning tasks but also exhibits strong generalizability on standard referring and open-vocabulary segmentation. We believe this work provides a robust foundation for future context-aware embodied agents in complex 3D environments.

\begin{acks}
This work was supported by the Natural Science Foundation of China (62531022) and the Guangdong Provincial Key Laboratory of Ultra High Definition Immersive Media Technology (Grant No. 2024B1212010006).
\end{acks}

\bibliographystyle{ACM-Reference-Format}
\bibliography{ref}

\clearpage
\appendix

\section{Additional Results}
\subsection{Additional Quantitative Results}
To comprehensively validate the reasoning capability of our method beyond overall performance, we conduct fine-grained quantitative evaluation on the Causal-LERF dataset, with results broken down by the four progressive reasoning dimensions defined in Section~\ref{sec:taxonomy}: Spatial Reasoning, Commonsense Reasoning, Affordance Reasoning, and Predictive and Counterfactual Reasoning. This fine-grained analysis verifies that our method excels at high-level complex reasoning tasks, rather than only achieving good performance on simple semantic matching subsets. Our method outperforms all baselines by a large margin across all four reasoning dimensions, achieving mIoU of 58.9\%, 42.9\%, 49.5\%, and 50.6\% on Spatial Reasoning, Commonsense Reasoning, Affordance Reasoning, and Predictive and Counterfactual Reasoning tasks, respectively.

Notably, our method achieves the most significant performance gain on the most challenging tasks with high reasoning complexity. For Predictive and Counterfactual Reasoning, the highest cognitive level requiring modeling of object state changes and counterfactual assumptions, our method reaches 50.6\% mIoU, outperforming the previous state-of-the-art ReferSplat (16.0\%) by a 34.6 absolute mIoU gain. For Spatial Reasoning, the core task for 3D scene understanding requiring parsing of geometric relations and 3D topologies, our method achieves 58.9\% mIoU, exceeding the previous best LUDVIG (23.4\%) by 35.5 absolute mIoU. These results demonstrate the superiority of our scene graph based framework in modeling complex spatial relationships and high-level semantic reasoning. On Affordance Reasoning and Commonsense Reasoning tasks, our method also consistently outperforms all baselines, with 49.5\% and 42.9\% mIoU respectively, validating the effectiveness of our multimodal node representation in aligning visual features, object attributes and functional semantics with natural language instructions.
\begin{table*}[htbp]
	\centering
	\caption{Fine-grained average mIoU results on the Causal-LERF dataset, broken down by the reasoning taxonomy defined in Section~\ref{sec:taxonomy}. The best result is bolded.}
		\vspace{-6pt}
	\label{tab:fine_grained_results}
	\resizebox{\linewidth}{!}{
		\begin{tabular}{lcccccc}
			\toprule
			Reasoning Dimension & OpenGaussian & Dr.Splat & InstanceGaussian & LUDVIG & ReferSplat & Ours \\
			\midrule
			Spatial Reasoning & 2.5 & 14.4 & 5.9 & 23.4 & 14.5 & \textbf{58.9} \\
			Commonsense Reasoning & 4.1 & 13.7 & 8.3 & 28.1 & 8.1 & \textbf{42.9} \\
			Affordance Reasoning & 3.2 & 9.9 & 2.9 & 34.3 & 7.3 & \textbf{49.5} \\
			Predictive and Counterfactual Reasoning & 1.8 & 5.6 & 2.1 & 10.8 & 16.0 & \textbf{50.6} \\
			\bottomrule
		\end{tabular}
	}
\end{table*}

\subsection{Additional Qualitative Results}

To further validate the generalization capability of our approach, we provide additional qualitative evaluations on the Ref-LERF and LERF datasets.

\begin{figure*}[t]
	\centering
	\includegraphics[width=\linewidth]{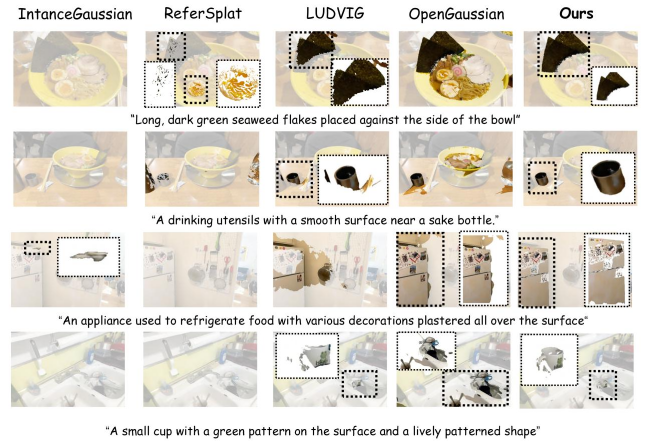}

	\caption{	\vspace{-6pt}Qualitative segmentation results on the Ref-LERF dataset.}

	\label{fig:f02}
\end{figure*}

\noindent \textbf{Results on Ref-LERF.} Figure~\ref{fig:f02} visualizes the 3D localization and segmentation performance on the Ref-LERF dataset. Compared with baselines (InstanceGaussian, ReferSplat, LUDVIG, OpenGaussian) that suffer from inaccurate positioning, blurred boundaries and severe background noise, our method accurately grounds natural language queries to target regions, yields sharper object boundaries and effectively suppresses artifacts. This demonstrates the superiority of our spatial weighting and contrastive optimization strategies for complex spatial reference tasks.

\begin{figure*}[t]
	\centering
	\includegraphics[width=\linewidth]{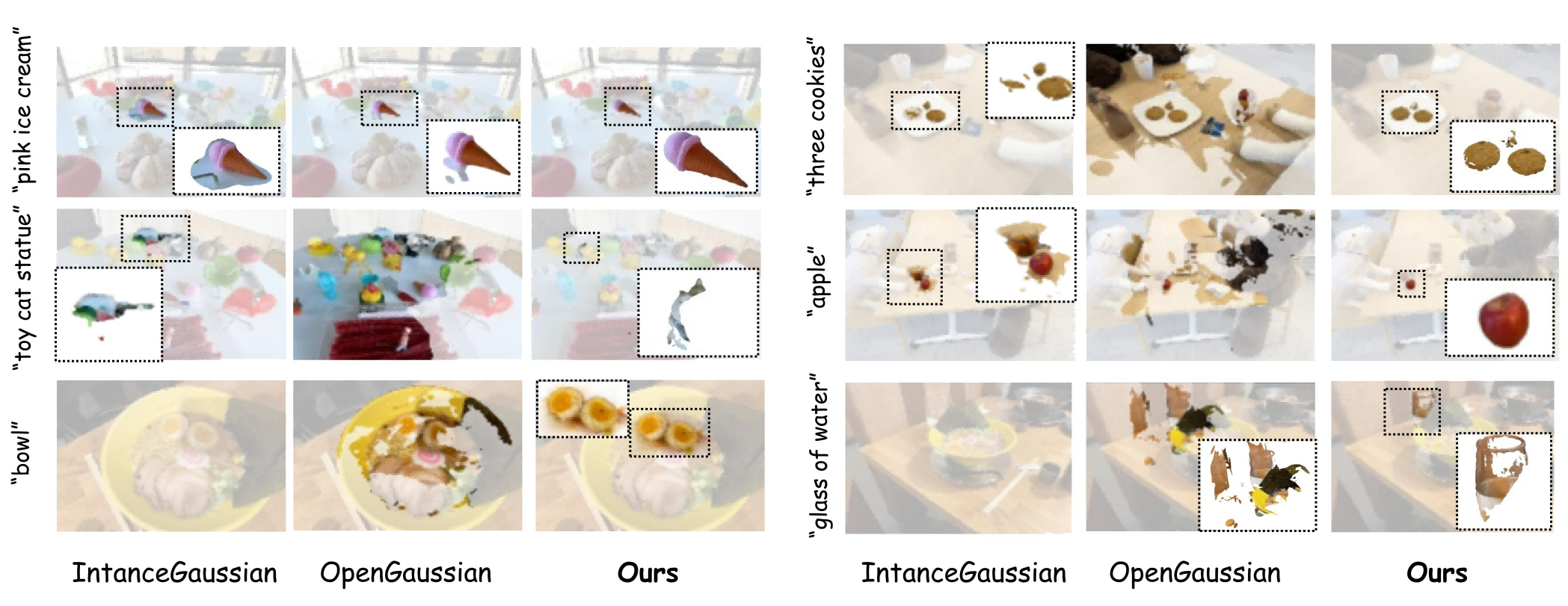}
	\caption{Qualitative search results on the LERF dataset.}
	\label{fig:f03}
\end{figure*}

\noindent \textbf{Results on LERF.} Figure~\ref{fig:f03} illustrates the visual search results on the LERF dataset. Our framework robustly identifies target objects across varying scales and complex indoor layouts, and successfully distinguishes queried instances from visually similar distractors. By contrast, baselines (InstanceGaussian, OpenGaussian) exhibit obvious defects including missed detection, semantic confusion and incomplete reconstruction. These results confirm that our method maintains strong semantic alignment and precise instance decoupling capabilities for arbitrary text queries.

\subsection{Generalization on Expanded Dataset}
To better assess broad generalization, we expanded our datasets using the same data generation and annotation methods described in the paper. As shown in the Table~\ref{tab:expanded_dataset}, this increases the total instructions from 231 to 2355. Evaluated on this larger scale, our approach consistently outperforms the state of the art baseline Ludvig in average accuracy on both datasets. These results validate our claims of zero-shot capabilities and broad generalization.

\subsection{Additional Ablation Studies}
We evaluate the effect of the relative displacement threshold $\tau$ along the principal axis. This hyperparameter controls the condition for establishing spatial edges between 3D entities in the scene graph. On the Causal-LERF dataset, we test three values and find that $\tau=0.3$ yields the best performance with an mIoU of 47.0\%. When $\tau$ is too small (0.1), the spatial constraint becomes too loose: the model creates false edges between weakly related entities, which introduces topological noise and reduces the mIoU to 42.8\%. Conversely, when $\tau$ is too large (0.6), the constraint becomes too strict: the model misses valid spatial relationships, leading to an incomplete scene graph and a lower mIoU of 45.1\%. Therefore, we choose $\tau=0.3$ as the optimal default value to maintain a balanced and accurate semantic scene graph.

\section{Analysis of Failure Cases}
\label{afc}
We conduct a systematic analysis of the failure cases of our method, and identify two core limitation-induced failure modes, which provide directions for further optimization of our framework.

\begin{figure}[t]
	\centering
	\includegraphics[width=\linewidth]{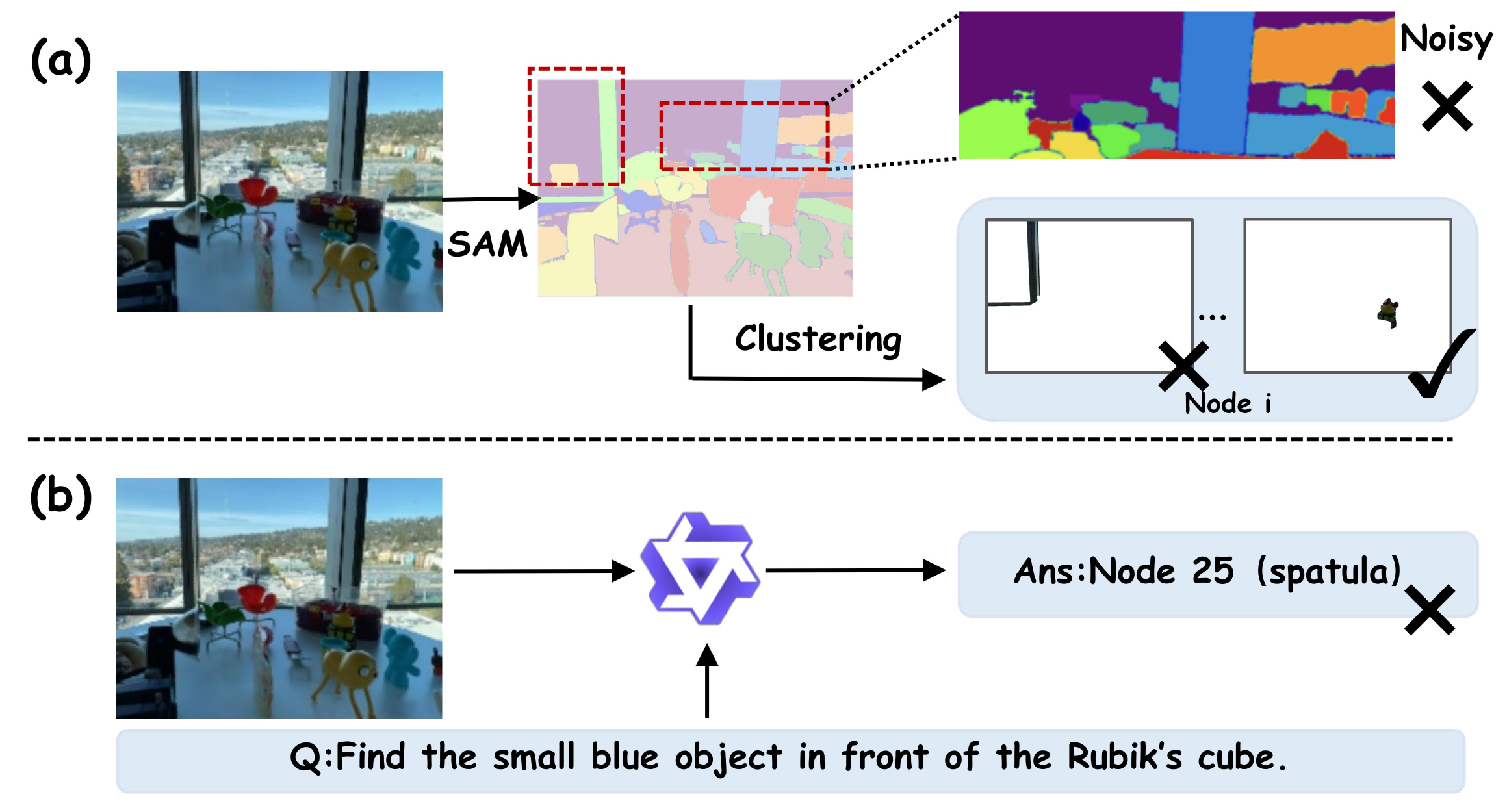}
	\caption{Core failure modes of our method. (a) Extreme segmentation noise causes 3D instance feature contamination and localization failure. (b) Ultra-complex spatial constraints cause VLM semantic and target node matching errors.}
	\label{fig:failure_case}
\end{figure}

\noindent \textbf{3D Instance Feature Contamination under Extreme Segmentation Noise}
Our method relies on multi-view 2D masks generated by SAM as the foundational input for semantic field construction. When the scene contains large-area textureless backgrounds or severely cluttered foreground occlusion, SAM may produce large-scale and persistent segmentation errors, such as massive fragmented invalid masks from background regions, and severe missing or broken masks of the target foreground. Although our spatial weighting and contrastive optimization modules can suppress partial noise, they cannot fully correct such extreme input errors. These invalid masks will be mixed into the feature clusters of target instances during the clustering process, contaminating the semantic features of 3D entities, and eventually leading to incorrect instance assignment and target localization failure, as shown in Fig.~\ref{fig:failure_case}(a).

\noindent \textbf{VLM Semantic Matching Failure under Ultra-complex Spatial Constraints}
Our method implements reasoning and matching for complex language queries based on the structured scene graph, which provides spatial and semantic cues for the VLM. However, when the language query contains multi-level nested spatial constraints, or there are a large number of interfering instances with highly similar visual features in the scene, the VLM may still make mistakes in the semantic parsing of the query and the topological search of the scene graph. For example, when the query contains nested descriptions of multiple groups of relative positional relationships, the VLM may incorrectly parse the core constraint conditions, resulting in wrong target node matching and failure to locate the correct 3D instance, as illustrated in Fig.~\ref{fig:failure_case}(b).

\begin{table*}[t]
	\centering
	\caption{Comparison of average mIoU on expanded Causal-LERF and Causal-ScanNet datasets.}
	\vspace{-6pt}
	\label{tab:expanded_dataset}
	\begin{tabular}{l*{6}{>{\centering\arraybackslash}p{0.13\linewidth}}}
		\toprule
		Method & Teatime & Ramen & Figurine & Waldo & CausalLERF\newline Avg & CausalScanNet \\
		\midrule
		Ludvig & 22.5 & \textbf{27.7} & 28.3 & 10.2 & 22.2 & 6.7 \\
		Ours & \textbf{56.1} & 26.7 & \textbf{39.4} & \textbf{39.8} & \textbf{40.5} & \textbf{14.3} \\
		\bottomrule
	\end{tabular}
\end{table*}

\begin{table}[t]
	\centering
	\caption{Ablation study on the relative distance threshold $\tau$ along the principal axis on the Causal-LERF dataset. We report the 2D mIoU (\%) metric.}
	\label{tab:ablation_threshold}
	\setlength{\tabcolsep}{24pt}
	\renewcommand{\arraystretch}{1.1}
	\begin{tabular}{lc}
		\toprule
		\textbf{Threshold $\tau$} & \textbf{mIoU $\uparrow$} \\
		\midrule
		0.1 & 42.8 \\
		0.3 (Ours) & \textbf{47.0} \\
		0.6 & 45.1 \\
		\bottomrule
	\end{tabular}
\end{table}

\section{Dataset Details}
We construct a testbed for 3D scene language understanding using two datasets, Causal-LERF and Causal-ScanNet. In total, they contain 231 natural language queries covering four core reasoning dimensions: spatial reasoning, commonsense reasoning, affordance reasoning, and predictive and counterfactual reasoning.

Causal-LERF consists of 158 queries, distributed as follows: 27 (17.1\%) on spatial reasoning, 79 (50.0\%) on commonsense reasoning, 25 (15.8\%) on affordance reasoning, and 27 (17.1\%) on predictive and counterfactual reasoning. Causal-ScanNet contains 73 queries, with the following distribution: 15 (20.5\%) on spatial reasoning, 33 (45.2\%) on commonsense reasoning, 12 (16.4\%) on affordance reasoning, and 13 (17.8\%) on predictive and counterfactual reasoning.

Across the entire dataset, commonsense reasoning queries are the most frequent, accounting for 48.5\% and serving as the primary reasoning type in 3D scene language understanding. Spatial reasoning, affordance reasoning, and predictive and counterfactual reasoning constitute 18.2\%, 16.0\%, and 17.3\%, respectively. This diverse distribution ensures coverage of core reasoning dimensions, enabling comprehensive evaluation of models' reasoning and understanding capabilities.

\begin{table}[t]
	\centering
	\caption{Runtime comparison (unit: second) on the Teatime scene with single RTX 4090D GPU.}
	\label{tab:runtime_compare}
	\resizebox{\linewidth}{!}{
		\begin{tabular}{lcccc}
			\toprule
			Method & Mask Ext. & Training & Processing & Total \\
			\midrule
			OpenGaussian & 5243 & 3993 & 1 & 9237 \\
			InstanceGaussian & 5243 & 8518 & 0 & 10333 \\
			ReferSplat & 1813 & 5117 & 13 & 6943 \\
			DrSplat & 5243 & 4572 & 3189 & 13004 \\
			Ours & \textbf{4354} & \textbf{5460} & \textbf{997} & \textbf{10811} \\
			\bottomrule
		\end{tabular}
	}
\end{table}
\section{Computational Efficiency}
Our pipeline consists of five stages: mask extraction, feature training, clustering, graph construction, and multimodal reasoning. We evaluate the processing time on the Teatime scene from the LERF dataset using a single RTX 4090D GPU. By decoupling perception from reasoning, our method enables higher-order reasoning while maintaining an overall runtime comparable to existing baselines. Regarding GPU memory, the initial feature processing stages require 16.5 GiB (peaking at 18.5 GiB). However, memory consumption drops significantly during the subsequent clustering, graph construction, and reasoning stages to 12.1 GiB, 2.6 GiB, and 3.7 GiB, respectively. This indicates a practical memory footprint for standard hardware, and the detailed runtime across all pipeline stages are summarized in Table~\ref{tab:runtime_compare}.

\end{document}